\pdfoutput=1

\documentclass[11pt]{article}

\usepackage{latex/acl}
\usepackage{graphicx}

\usepackage{times}
\usepackage{latexsym}
\usepackage{makecell}
\usepackage[T1]{fontenc}

\usepackage[utf8]{inputenc}

\usepackage{microtype}

\usepackage{inconsolata}
\usepackage{url}
\usepackage{hyperref}
\usepackage{amsmath}
\usepackage{multirow}
\usepackage{makecell}
\title{Phased Instruction Fine-Tuning for Large Language Models}

\author{Wei Pang$^\dag$ \\ \texttt{pangweitf@bupt.cn}  
  \And
  Chuan Zhou$^\dag$ \\ Peking University 
  \AND 
  Xiao-Hua Zhou$^\ast$ \\ Peking University
  \And
  Xiaojie Wang$^\ast$ \\ Beijing University of Posts and Telecommunications \\
\\}

\begin{document}
\maketitle

\def\thefootnote{$^\dag$}\footnotetext{Equal contribution}
\def\thefootnote{$^\ast$}\footnotetext{Corresponding author}

\begin{abstract} 

Instruction Fine-Tuning enhances pre-trained language models from basic next-word prediction to complex instruction-following. However, existing One-off Instruction Fine-Tuning (One-off IFT) method, applied on a diverse instruction, may not effectively boost models' adherence to instructions due to the simultaneous handling of varying instruction complexities. To improve this, Phased Instruction Fine-Tuning (Phased IFT) is proposed, based on the idea that learning to follow instructions is a gradual process. It assesses instruction difficulty using GPT-4, divides the instruction data into subsets of increasing difficulty, and uptrains the model sequentially on these subsets. Experiments with Llama-2 7B/13B/70B, Llama3 8/70B and Mistral-7B models using Alpaca data show that Phased IFT significantly outperforms One-off IFT, supporting the progressive alignment hypothesis and providing a simple and efficient way to enhance large language models. Codes and datasets from our experiments are freely available at \href{https://github.com/xubuvd/PhasedSFT}{https://github.com/xubuvd/PhasedSFT}.

\end{abstract}

\section{Introduction}

\begin{figure}[!htb]
\centering 
\includegraphics[width=0.5\textwidth,scale=0.8, clip=true, keepaspectratio]{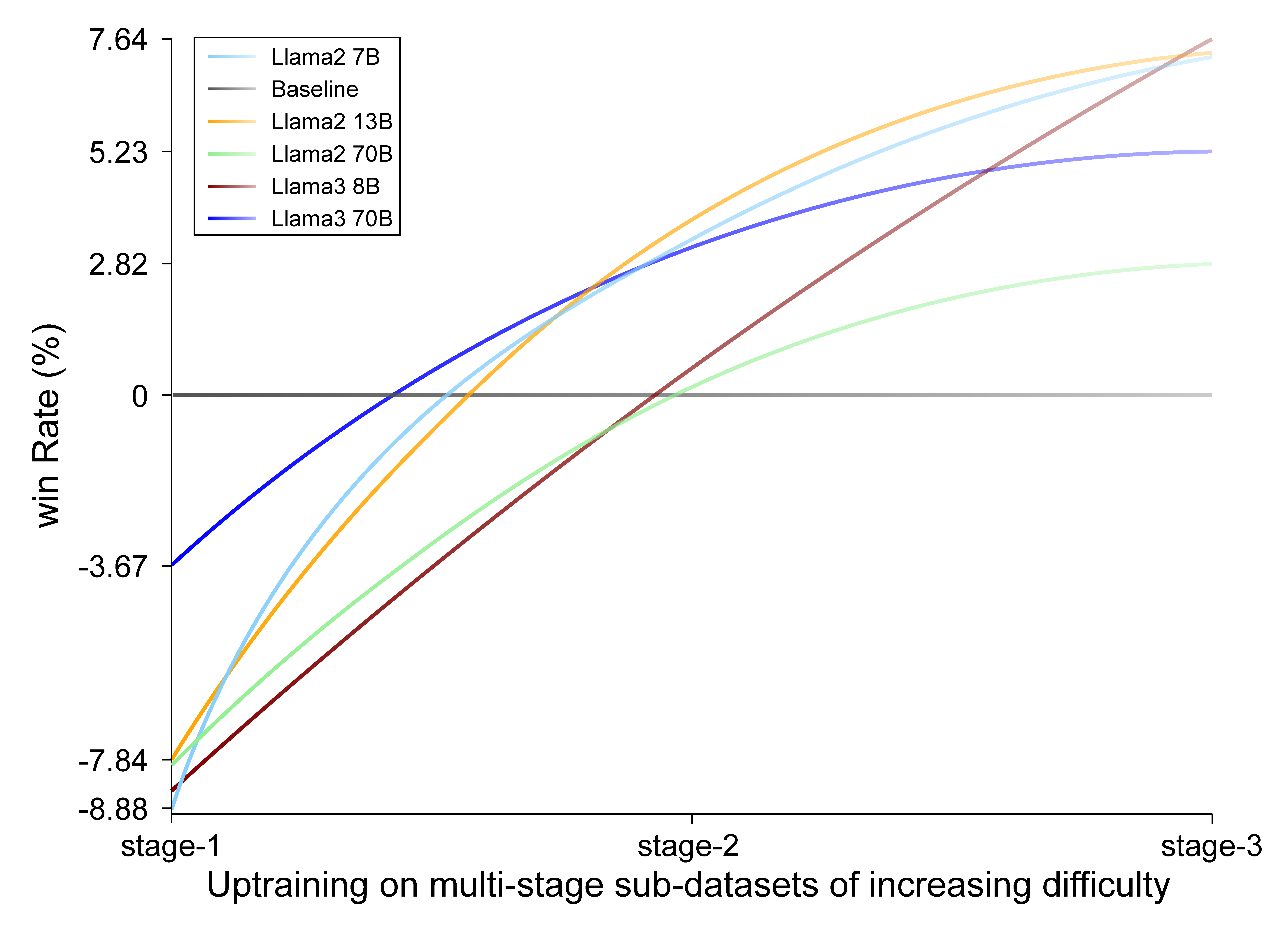}
\caption{In the context of increasing difficulty multi-stage sub-datasets, the trend of win rate growth for uptraining (Phased IFT) compared to One-off on the original dataset is observed. The gray horizontal line represents the performance baseline of One-off.}

\label{fig_first}
\end{figure}

Instruction fine-tuning (IFT) \cite{InstructGPT,Flan}, involving training on instruction dataset using standard supervised fine-tuning method, aligns pre-trained language models to users’s intent and has been proven as an effective alignment method to enhance their ability to follow instructions. Large language models (LLMs) are pre-trained on raw text data using maximum likelihood estimation, equipping them with the basic ability to predict the next word \cite{LIMA,zhao2023survey}. However, a gap exists between this ability and following user intentions \cite{LaMDA,InstructGPT}. To bridge this gap and enable models to complete human end tasks, various instruction fine-tuning strategies have been proposed, including SFT \cite{InstructGPT}, LIMA \cite{LIMA}, Alpaca \cite{Alpaca,AlpacaFarm}, Alpagasus \cite{Alpagasus}, CoT \cite{CoT}, Superfiltering \cite{li2024superfiltering} and Self-instruct \cite{SELF-INSTRUCT}. Among these, SFT and LIMA employ human-written instruction data for fine-tuning, while strategies like Self-instruct, Alpaca, Alpagasus utilize ChatGPT as a teacher to automatically generate extensive instruction datasets. Their training approach, a one-off IFT on the whole instruction data without differentiating the difficulty levels of instructions, lacks efficiency in enhancing the instruction-following capability of pre-trained models.

An instruction sample comprises a triplet of an instruction, an optional input, and an output \cite{SELF-INSTRUCT}. The instruction describes the task, the input serves as an additional context to the instruction, and the output is the answer following the instruction. Diversity \cite{AlpacaFarm,SELF-INSTRUCT,LIMA} is a crucial aspect of a high-quality instruction dataset, indicating that such datasets are typically extensive, encompassing a wide range of tasks and examples representing different levels of instruction-following difficulty. For instance, tasks like mathematical problem-solving, code writing, entity extraction, and copy generation each present varying levels of difficulty. Even within the same category of instructions, the difficulty can vary with the length of the input (refer to Table~\ref{tab_gpt4} for details). In the realm of entity extraction, the difficulty differs between extracting place names and identifying product name fragments in e-commerce \cite{EcomGPT}.

However, previous instruction fine-tuning methods have treated large-scale instruction dataset uniformly, feeding them to the pre-trained model for one-time alignment training without differentiating the role of instructions of varying difficulties in the fine-tuning process. This approach overlooks the nuanced differences in instruction complexities and potentially limits the efficiency of model training.

To address the aforementioned issue, this paper introduces an instruction difficulty scoring mechanism. Specifically, the difficulty of an instruction is defined by the complexity of the instruction itself, the input, and the challenges involved in generating the output. We employ the strongest available models (such as GPT-4 \cite{GPT-4}) as a teacher to score the difficulty of each instruction and input, as well as the challenge in generating the output, on a scale from 1 to 5, where higher scores indicate greater difficulty. Further, we plot probability density curve of the instruction difficulty as a heuristic guide \cite{campello2020density}, coupled with expert judgment to select thresholds for difficulty scores. This approach enabled us to segment the instruction dataset into multi-stages sub-datasets, forming a sequence with increasing difficulty levels. Building upon this, we further propose the {\bfseries Progressive Alignment Hypothesis}: Aligning a pre-trained model's existing ability to predict the next word with the capability to generate content following human intent is a gradual learning process, progressively attaining alignment with human intent.

Based on the hypothesis, we develop a Phased Instruction Fine-Tuning (Phased IFT) method, diverging from the traditional One-off IFT approach. Phased IFT represents an effective sequential uptraining framework, which entails progressively training on sub-instruction datasets of incremental difficulty. Initially, training commences on an easy sub-dataset with the standard supervised loss. Following training completion, the model checkpoint is saved and subsequently utilized to extend supervised training to a marginally more complex sub-dataset. This iterative process persists until the most challenging sub-dataset is addressed.

We conducted extensive experiments using two leading open-source pre-trained models, Llama2 7/13/70B \cite{LLaMA2}, Llama3 8/70B, and Mistral 7B \cite{Mistral7B}, on two widely-used instruction datasets, Alpaca and its refined version AlpacaClean. To evaluate the training effectiveness, we utilized 6 benchmarks, with GPT-4 serving as the judge to measure the performance improvements of Phased IFT over One-off IFT in terms of win rate. Specially, Figure~\ref{fig_first} illustrates the process wherein Phased IFT's win rate progressively surpasses that of One-off IFT on the three-stage difficulty-increasing sub-datasets. It is evident that the three Llama2 models and the two Llama3 models exhibit a consistent trend of increasing win rates with the progression of uptraining on multi-stage sub-datasets, indicating the effectiveness of the progressive alignment hypothesis. Furthermore, we conducted three sets of experiments and an ablation study to confirm the effectiveness of Phased IFT. The main contributions of this paper include:

\begin{itemize}

\item We propose a phased instruction fine-tuning (Phased IFT) method, utilizing GPT-4 for scoring the difficulty of instructions and subsequently dividing the instruction dataset into a sequence of multi-stages sub-datasets with increasing difficulty levels. This approach employs uptraining on the sequence of multi-stages sub-datasets.

\item We introduce the Progressive Alignment Hypothesis, which posits that aligning the generative capabilities of pre-trained models with human intent is a gradual process, rather than being achieved through a one-time fine-tuning on an instruction dataset. This hypothesis is supported by extensive experimental validation.

\item Extensive experiments have demonstrated that our proposed Phased IFT is more effective than One-off IFT. It has significantly outperformed One-off IFT across 6 benchmarks. Additionally, a series of ablation experiments have validated the effectiveness of the Progressive Alignment Hypothesis.
\end{itemize}

\section{Related Work}


{\bf Open-source Instruction Dataset}
The Alpaca dataset \cite{Alpaca} aims to improve instruction-following ability of LLMs, consisting of 52K instructions, generated using OpenAI's text-davincic-003. It has improved the self-instruction framework \cite{SELF-INSTRUCT}, leading to a greater diversity in the Alpaca dataset compared to the Self-Instruction dataset. Alpaca-cleaned \cite{alpaca-cleaned} is a quality-enhanced version of the original Alpaca dataset, which addresses hallucinations, empty outputs, and incorrect answers, thereby improving the overall quality of the dataset. In addition, LIMA \cite{LIMA} has also released a carefully curated dataset, including 1000 instructions, which encompasses 750 popular queries and responses from community forums such as Stack Exchange and wikiHow, as well as 250 meticulously crafted entries.

{\bf Instruction Fine-tuning Strategy} 
The prevalent method of instruction fine-tuning, employed in models such as LIMA, Alpaca, Alpagasus and Superfiltering \cite{Quantity,li2024superfiltering}, involves a one-off instruction fine-tuning (One-off IFT) on the whole instruction dataset characterized by their diversity and high quality. This approach, while emphasizing dataset quality and diversity \cite{SELF-INSTRUCT}, neglects the inherent complexity of the instruction sets. Consequently, this one-off IFT fails to adequately equip models with the nuanced capability to comprehend and execute a wide array of instructions. Differs from the one-off IFT approach, our proposed phased IFT method initially segments the instruction dataset into multiple sub-datasets, arranged in a sequence from simple to complex in terms of instruction difficulty. Continuous uptraining is then executed across this sequence.

\section{Method}

\begin{figure*}[!htb]
\centering 
\includegraphics[width=0.95\textwidth,scale=0.8, clip=true, keepaspectratio]{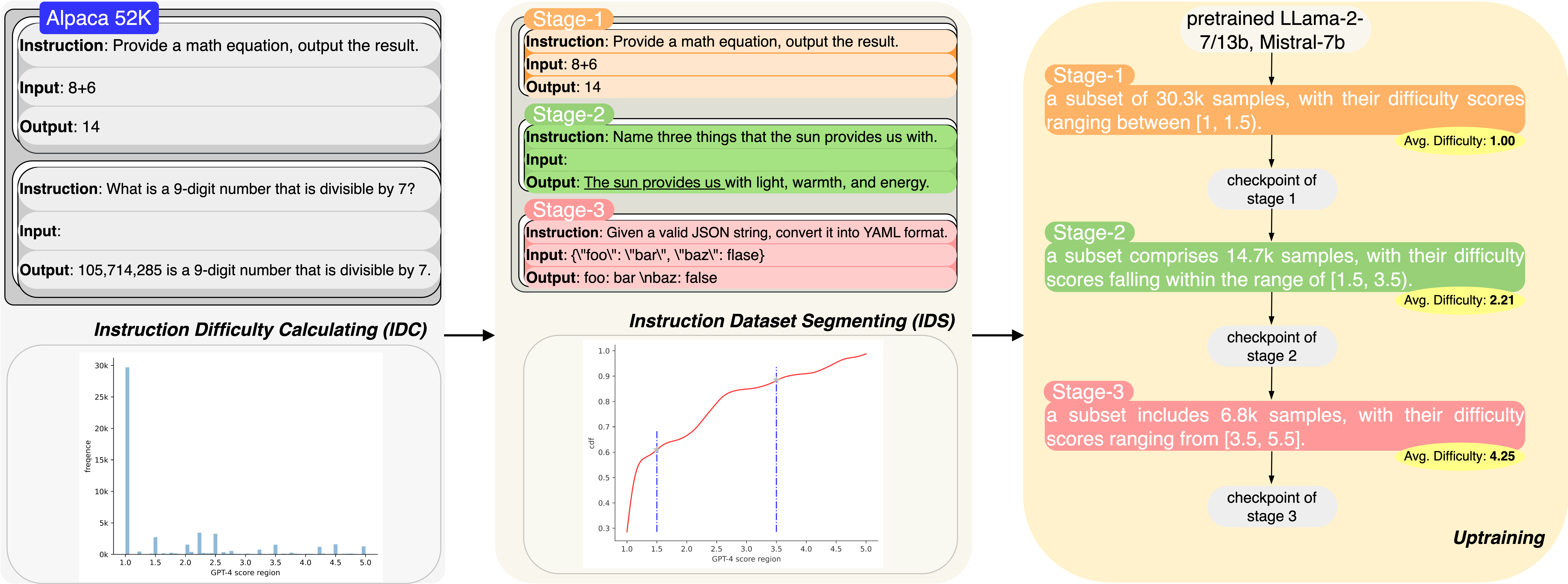}
\caption{Overview of the proposed Phased Instruction Fine-Tuning (Phased IFT).}
\label{fig_overview}
\end{figure*}

Figure~\ref{fig_overview} presents the overall pipeline of our proposed Phased Instruction Fine-Tuning (Phased IFT), encompassing three key components: Instruction Difficulty Calculating (IDC), Instruction Dataset Segmenting (IDS), and Uptraining. Where, IDC prompts GPT-4 to allocate a difficulty rating to each instruction. Following this, IDS converts the instructions's score into a cumulative probability distribution, which facilitates the division of the dataset into subsets categorized by difficulty level, and segments the whole instruction data into several stages. Finally, across this stages, Uptraining is performed, beginning with those classified as less challenging. Detailed elaborations on the specifics of each component are provided in the subsequent sections of this chapter.

\begin{figure*}[!htb]
\centering 
\includegraphics[width=0.95\textwidth,scale=0.8, height=6cm, clip=true, keepaspectratio]{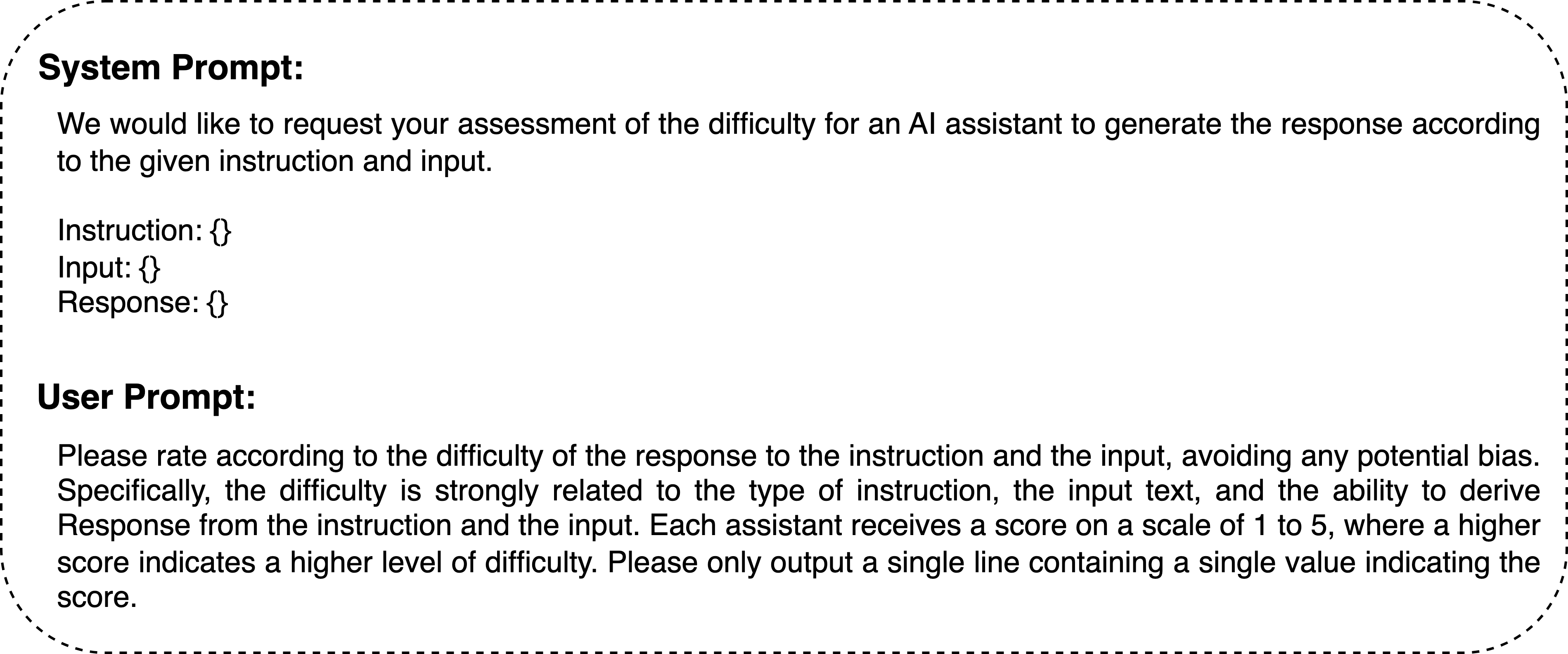}
\caption{A prompt to ChatGPT-4 for scoring instruction difficulty.}
\label{fig_prompt}
\end{figure*}

\noindent {\bf Instruction Difficulty Calculating (IDC)}
Following Alpagasus's prompt \cite{Alpagasus}, we also design a GPT-4-based scoring prompt to assess the difficulty of each triplet of (instruction, input, output) within the 52K Alpaca dataset, using a scale of 1 to 5, with higher scores indicating greater difficulty. This prompt is detailed in Figure~\ref{fig_prompt}, where each triplet is evaluated in two dimensions. Firstly, we consider the intrinsic difficulty of the instruction itself, reflecting the complexity inherent to the task described. Secondly, we evaluate the challenge involved in transforming the given instruction to the expected output, thereby estimating the difficulty from instruction to output. This dual-dimensional scoring approach allows for a nuanced understanding of task complexity within the dataset.

The difficulty score histogram of the 52K Alpaca dataset is illustrated as shown in the left of Figure~\ref{fig_overview}. Another interesting finding is that a majority of instructions in the 52K Alpaca are categorized as low difficulty, with approximately 63.5\% scoring below 1.5. In contrast, within the Alpaca-clean dataset, instructions with a difficulty score lower than 1.5 account for 52.3\%; in the LIMA dataset, the proportion of instructions with a difficulty score below 1.5 is merely 9.9\%, while those scoring above 4.5 account for a significant 58.3\%. Details of their comparison can be found in Appendix~\ref{sec:appendix_sec3}and~\ref{sec:appendix_sec1}. This highlights the feasibility of employing GPT-4 to assess the difficulty level of instructions.

\noindent {\bf Instruction Dataset Segmenting (IDS)}

Our goal is to divide the instruction dataset into multi-stages sub-datasets based on difficulty scores, while ensuring that the number of samples in each sub-dataset remains relatively balanced. To obtain heuristic information from the scored, discrete dataset, we plot Gaussian kernel density curve on the scored dataset. This curve represents the cumulative probability density as the difficulty score increases from the minimum to the current score, as depicted in the middle part of Figure~\ref{fig_overview}. Utilizing human experience, we select threshold scores, such as the 1.5 and 3.5 marked by blue vertical lines. These thresholds segment the whole 52K Alpaca instruction dataset into three stages sub-datasets. 

The first stage (stage 1), covering difficulty scores from [1, 1.5), encompasses 30.3k samples; the second stage (stage 2), spanning scores from [1.5, 3.5), accounts for 14.7k samples; and the third stage (stage 3), with scores within [3.5, 5], comprises 6.8k samples. These three stages sub-datasets form a sequence of sub-datasets with increasing difficulty scores, which are used for uptraining in the next phase.

\noindent {\bf Uptraining}
In the post-IDS phase, uptraining is performed across three sequential stages, initiating with fine-tuning at the first stage and progressively advancing through the second and third stages.  Throughout this process, hyperparameters such as learning rate, batch size, epoch count, and warmup ratio are consistently applied.  A standard supervised loss function is used in the uptraining.  Specifically, the triplet of (instruction, input, output) is merged into a text string for training, wherein the model predicts each token.  Nonetheless, loss calculation is confined to the output part, with loss from the instruction and input parts masked.

\section{Experiments}
This paper aims to validate the progressive alignment hypothesis by comparing the efficacy of two instruction training methodologies: One-off Instruction Fine-Tuning (One-off IFT) and our proposed Phased Instruction Fine-Tuning (Phased IFT), through three experiments and ablation studies. Experiment 1 demonstrates the superior performance of Phased IFT over One-off IFT across two instruction datasets, establishing its effectiveness. Experiment 2 evaluates the effectiveness of stratifying instruction datasets by difficulty versus arbitrary segmentation, aiming to attribute Phased IFT's success to difficulty-stratified instruction data categorization rather than to the phased training approach. Experiment 3 conducts a full permutation analysis of Phased IFT to demonstrate that its success is specifically due to uptraining in a sequence of increasing difficulty, with alternative sequencing failing to produce comparable outcomes.

\noindent {\bf Datasets and Evaluation}
We conducted extensive experiments on the standard Alpaca 52K instruction data and its refined version, Alpaca-cleaned, using 6 recent LLMs: Llama-2 7/13/70B, Llama-3 8/70B and Mistral-7B. These models were evaluated across 6 benchmark datasets: Self-Instruction \cite{SELF-INSTRUCT} of size 252, WizardLM \cite{WizardLM} 218, Koala \cite{Koala} 180, Vicuna \cite{Vicuna} 80, OASST 188 \cite{kopf2023openassistant}, and Anthropic of size 129 \cite{Anthropic}.

Following the evaluation framework established by Alpagasus \cite{Alpagasus}, we prompted GPT-4 to assess Win-Tie-Lose outcomes and compute win rates across the aforementioned 6 benchmarks. Subsequently, a weighted average win rate metric is calculated based on these 6 benchmark datasets, referred to as Avg. win rate. Using identical instruction and input, GPT-4 acts as an auto-grader \cite{zheng2023judging,chia2023instructeval} to score the quality of output from both One-off IFT and Phased IFT methods on a scale of 1 to 10 (see prompt in Appendix~\ref{sec:appendix_sec2}). To ensure fairness, each instruction is scored twice, with Phased IFT's output positioned before and after One-off IFT's output. An instruction's overall outcome is determined based on these evaluations.
\begin{itemize}
\item Win: win twice or win once and tie once; 
\item Tie: tie twice or win once and loses once;
\item Lose: loses twice or loses once and tie once.
\end{itemize}
The win rate calculation formula is as follows:
\begin{align*} \label{equ_winrate}
\mathrm{WinRate} = \frac{\#win + \#tie/2}{\#win + \#tie + \#lose} - 0.5,
\end{align*}
where \# means the corresponding number of samples.

\begin{table*}[!htb]
\centering
\caption{Detailed statistics in each stage of Alpaca and Alpaca-cleaned instruction data, including the number of samples (K), average difficulty per sample, and average token length per sample.} \label{tab_gpt4}
\setlength{\tabcolsep}{1.8mm}{
\begin{tabular}{lccccccccccc}
\hline
\multirow{2}{*}{\bfseries Data}&\multicolumn{3} {c}{\bfseries Number (K)} && \multicolumn{3} {c}{\bfseries Avg. Difficulty} && \multicolumn{3} {c}{\bfseries Avg. Token Length}\\ \cline{2-4} \cline{6-8} \cline{10-12}
                                &stage1&stage2&stage3 && stage1 & stage2 & stage3 &&stage1 & stage2 & stage3\\ 
Alpaca&30.3&14.7&6.8&&1.00&2.21&4.25&&63.16&106.83&134.40\\ 
Alpaca-cleaned&23.9&16.4&11.4&&1.00&2.22&4.25&&113.42&200.99&301.60\\ \hline
\end{tabular}}
\end{table*}

\noindent {\bf Stratifying Alpaca by Difficulty Score}
The instruction difficulty within the Alpaca and Alpaca-cleaned datasets is quantitatively assessed by GPT-4, assigning scores from 1 to 5, with higher scores indicating increased complexity. Using density curves (e.g., Figure~\ref{fig_overview} for Alpaca and Figure~\ref{fig_appendix_AlpacaCleaned} for Alpaca-cleaned), we adopt threshold values of 1.5 and 3.5 for both datasets, segmenting them into three stages sub-datasets, referred to as Alpaca-3-stages and Alpaca-cleaned-3-stages.

For ablation experiments, we construct randomly sampled 3-stages datasets, with each stage containing the same number of instructions as the difficulty-segmented subset. These are denoted as Alpaca-rand-3-stages and Alpaca-cleaned-rand-3-stages. Detailed statistics are provided in Table~\ref{tab_gpt4}.

\noindent {\bf Implementations}
In our training, we employed Huggingface's Trainer coupled with DeepSpeed's Zero3 for data-parallel, full-parameter fine-tuning. Consistently, a learning rate of 5e-6 and a 2 epoch duration were utilized across experiments. We set a per-device batch size of 4 for 70B model and 16 for other models, a gradient accumulation step of 1, with a weight decay of 0.1 and a warmup ratio of 0.1. Additionally, a cosine annealing learning rate scheduler was employed.

\subsection {Experiment 1: compares the win rates of Phased IFT and One-off IFT}

\begin{table*}[!htb]
\centering 
\caption{Comparisons of win rates of Phased IFT against One-off IFT. The changes in win rates are represented by percentage differences, where a "+" indicates an increase and "-" denotes a decrease in percentage.} \label{tab_Exp1}
\setlength{\tabcolsep}{1.2mm}{
\begin{tabular}{l|cccccc|c|r}
\hline 
\multirow{1}{*}{Base Models}& {Self-Instruct} &{WizardLM}& {Koala} & {Vicuna} & {OASST} & {Anthropic}&Avg. win-rate&{Data}\\ \cline{1-9}
Llama-2 7B&+4.17&+6.42&+5.56&+6.88&+9.31&+14.34&+7.26&\multirow{6}{*}{Alpaca}\\
Mistral 7B&-0.60&+5.96&+6.11&+16.25&+7.98&+12.02&+6.30&\\
Llama-2 13B&+1.98&+2.75&+7.78&+11.88&+11.44&+16.28&+7.35&\\
Llama-2 70B&-3.04&+2.57&+7.27&+13.46&+4.10&+0.00&+2.82&\\
Llama-3 8B&+3.05&+9.67&+9.52&+7.50&+6.32&+12.60&+7.64&\\
Llama-3 70B&+1.65&+10.10&+5.23&+5.84&+2.34&+7.87&+5.23&\\ \hline
Llama-2 7B&+4.76&+1.15&+4.44&+2.50&+5.85&+1.16&+3.53&\multirow{5}{*}{\makecell[r]{Alpaca-\\cleaned}}\\
Llama-2 13B&+4.56&+2.98&+5.83&+15.62&+6.91&+10.85&+6.49&\\
Llama-2 70B&+0.40&+6.42&+8.33&+12.50&+10.90&+14.73&+7.59&\\
Llama-3 8B&+5.75&+3.90&+1.15&+0.62&+4.79&+5.43&+3.97&\\ \cline{1-9}
\hline
\end{tabular}}
\end{table*}

Table~\ref{tab_Exp1} lists the win rate results of Phased IFT compared to the One-off IFT method, demonstrating that our proposed Phased IFT significantly outperforms the One-off IFT method across six benchmarks on two training datasets.

The first part of Table~\ref{tab_Exp1} compares the results of the two fine-tuning methods on Alpaca. Phased IFT was uptrained on Alpaca-3-stages, while One-off IFT was trained once on the original Alpaca. Using six base models, Phased IFT exceeded One-off IFT in the Avg. win-rate metric. Specifically, compared to the One-off IFT method trained once on the original Alpaca dataset, the Llama-2 7B using the Phased IFT fine-tuning method on Alpaca-3-stages showed a minimum win rate increase of +4.17 on Self-Instruct and a maximum of +14.34 on Anthropic, with an average win rate of +7.26. The Mistral 7B also achieved an Avg. win rate of +6.30; Llama 2 13B reached an average win rate of +7.35, with a win rate of +16.28 on Anthropic; Llama 3 8B and 70B both achieved an average win rate increase of over +5.0.

The second part of Table~\ref{tab_Exp1} compares the win rates of five base models using the two fine-tuning methods on Alpaca-cleaned. Specifically, the three Llama2 series models achieved average win rates of +3.53, +6.49, and +7.59, respectively, while Llama3 8B achieved an average win rate of +3.97. It can be observed that the average win rates of Llama2 7B/13B and Llama3 8B on Alpaca-cleaned are lower than their results on Alpaca, while Llama2 70B's average win rate increased from +2.82 on Alpaca to +7.59 on Alpaca-cleaned. Due to time constraints, experiments on Llama3 70B have not yet been completed.

The results of Experiment 1 indicate that the Phased IFT fine-tuning method is indeed more effective than One-off IFT, achieving significantly higher win rates on most of the six benchmarks, with an overall average win rate significantly higher than that of One-off IFT.

\subsection {Experiment 2: Comparison of win rate between difficulty-stratified 3-stages and randomly sampled 3-stages} 

\begin{table*}[!htb]
\centering
\caption{Part \#1 assesses win rates of Phased IFT on Alpaca-rand-3-stages against One-off IFT on the original Alpaca data. Part \#2 evaluates win rates of Phased IFT on Alpaca-cleaned-rand-3-stages over One-off IFT on the original Alpaca-cleaned data. Part \#3 measures win rates of Phased IFT on Alpaca-3-stages compared to Phased IFT on Alpaca-rand-3-stages.} \label{tab_Exp2}
\setlength{\tabcolsep}{1.5mm}{
\begin{tabular}{l|cccccc|c|c}
\hline
\multirow{1}{*}{Base Models}& {Self-Instruct} &{WizardLM}& {Koala} & {Vicuna} & {OASST} & {Anthropic}&Avg. win-rate&{Part}\\ \cline{1-9}
Llama-2 7B&-0.60&+2.98&-6.61&+3.12&-1.33&+1.94&-0.42&\multirow{5}{*}{\#1}\\
Llama-2 13B&-1.98&-2.29&-0.29&-6.88&-2.93&-0.39&-2.10&\\
Llama-2 70B&-10.32&-5.96&-6.90&-3.12&-11.97&-5.43&-7.96&\\
Llama-3 8B&-7.74&-1.38&-1.15&+5.00&+0.00&-2.33&-2.25&\\
Llama-3 70B&-10.00&-3.63&-4.44&-5.63&-6.89&-2.73&-5.92&\\ \cline{1-9}
Llama-2 7B&-1.34&-5.07&-3.85&+8.12&-1.34&-2.76&-2.00&\multirow{3}{*}{\#2}\\
Llama-2 13B&+0.58&-0.69&+1.26&+1.88&+2.97&-0.39&0.84&\\
Llama-3 8B&+2.80&+0.69&-6.94&+2.50&+4.30&-6.59&-0.22&\\ \cline{1-9}
Llama-2 7B&+2.38&+3.90&+11.78&+11.25&+2.39&+8.53&+5.74&\multirow{5}{*}{\#3}\\
Llama-2 13B&+6.94&+2.98&+6.32&+16.25&+11.70&+12.79&+8.29&\\
Llama-2 70B&+8.93&+5.50&+10.34&+19.38&+11.97&+8.14&+9.71&\\
Llama-3 8B&+2.18&+0.69&+13.79&+2.50&+8.78&+9.30&+5.95&\\
Llama-3 70B&+8.07&+14.10&+12.50&+17.12&+8.63&+5.96&+10.62&\\ \cline{1-9}
\hline
\end{tabular}}
\end{table*}
\begin{table*}[!htb]
\centering
\caption{Win rates of all permutations of Phased IFT compared to One-off IFT on the original Alpaca data.} \label{tab_Exp3}
\setlength{\tabcolsep}{0.8mm}{
\begin{tabular}{l|cccccc|c|r}
\hline
\multirow{1}{*}{Permutations}& {Self-Instruct} &{WizardLM}& {Koala} & {Vicuna} & {OASST} & {Anthropic}&Avg. win-rate&{Base Models}\\ \cline{1-9}
{\bfseries 1-2-3}&+4.17&+6.42&+5.56&+6.88&+9.31&+14.34&+7.26&\multirow{6}{*}{Llama2 7B}\\
2-1-3&+1.39&+6.65&+2.30&+6.88&+4.79&+8.91&+4.59&\\
3-1-2&+0.80&+0.94&-2.35&+6.88&+3.83&+8.66&+2.26&\\
1-3-2&+3.57&-1.15&-1.15&+1.88&+2.66&+8.91&+2.14&\\
2-3-1&-2.61&-6.81&-7.65&-5.70&-4.12&-0.39&-4.5846&\\
3-2-1&-7.23&-3.18&-9.09&+1.54&-5.98&+2.73&-4.5847&\\ \hline
{\bfseries 1-2-3}&+1.98&+2.75&+7.78&+11.88&+11.44&+16.28&+7.35&\multirow{6}{*}{Llama2 13B}\\
2-1-3&+2.33&+1.63&+8.82&+6.39&+6.41&+12.29&+5.57&\\
3-1-2&-0.61&+0.70&+4.39&+5.62&+2.75&+8.40&+2.71&\\
1-3-2&-0.40&+2.14&+6.17&+1.25&+4.72&+9.60&+3.53&\\
2-3-1&-6.83&-10.56&-8.33&-8.12&-3.57&+0.78&-6.44&\\
3-2-1&-5.42&-8.22&-7.69&-8.23&-10.50&-1.97&-7.09&\\ \hline
{\bfseries 1-2-3}&+3.05&+9.67&+9.52&+7.50&+6.32&+12.60&+7.64&\multirow{6}{*}{Llama3 8B}\\
2-1-3&-2.48&+3.61&+4.94&+3.38&+6.21&+5.24&+3.02&\\
3-1-2&+0.82&+2.12&+0.90&+3.80&+2.53&+6.35&+2.32&\\
1-3-2&-0.41&+0.00&+0.31&-1.97&+5.40&+5.16&+1.40&\\
2-3-1&-11.27&-14.44&-11.70&-14.86&-16.44&-14.22&-13.57&\\
3-2-1&-11.88&-9.39&-5.00&-5.77&-9.23&-3.97&-8.26&\\ \hline
\end{tabular}}
\end{table*}
\begin{table*}[!htb]
\centering
\caption{Ablation studies of various methods on Alpaca-3-stages instruction data. Each entry denotes the win rate of Phased IFT against One-off IFT on the original Alpaca data.} \label{tab_Exp4}
\setlength{\tabcolsep}{0.8mm}{
\begin{tabular}{l|c|cccccc|c|r}
\hline
\multicolumn{2}{c|}{Method}&{Self-Instruct} &{WizardLM}& {Koala} & {Vicuna} & {OASST} & {Anthropic}&\makecell[c]{Avg. \\win-rate}&{Models}\\ \cline{1-10}
\multirow{3}{*}{Phased IFT}&stage1&-8.53&-10.32&-15.00&-6.25&-10.11&-4.26&-9.59&\multirow{9}{*}{\makecell[r]{Llama2\\7B}}\\
&stage2&+3.57&+2.98&-0.83&+5.00&+3.99&+7.36&+3.34&\\
&stage3&+4.17&+6.42&+5.56&+6.88&+9.31&+14.34&+7.26&\\ \cline{1-9}
\multicolumn{2}{l|}{IFT on stage2}&-0.20&+2.52&-0.83&+8.75&+0.53&+0.39&+1.14&\\ 
\multicolumn{2}{l|}{IFT on stage3}&+0.99&+6.88&+1.67&+16.25&+0.00&+10.08&+4.44&\\
\multicolumn{2}{l|}{IFT on mixed stage1\&2}&+2.78&+1.83&-8.33&-5.00&-4.79&+0.00&-1.62&\\
\multicolumn{2}{l|}{IFT on mixed stage1\&3}&-2.58&+0.00&-2.50&+3.12&-3.46&-3.88&-1.91&\\
\multicolumn{2}{l|}{IFT on mixed stage2\&3}&+0.40&+1.38&+4.17&+8.12&+1.33&+12.79&+3.53&\\ \hline \hline
\multirow{3}{*}{Phased IFT}&stage1&-6.55&-10.09&-13.61&-6.88&-6.65&-0.78&-7.83&\multirow{9}{*}{\makecell[r]{Llama2\\13B}}\\
&stage2&+0.40&-0.23&+4.44&+3.75&+3.72&+16.28&+3.77&\\
&stage3&+1.98&+2.75&+7.78&+11.88&+11.44&+16.28&+7.35&\\ \cline{1-9}
\multicolumn{2}{l|}{IFT on stage2}&-2.40&-0.47&+4.09&+3.80&+6.87&+9.06&+2.66&\\ 
\multicolumn{2}{l|}{IFT on stage3}&+4.56&+7.26&+4.11&+10.12&+3.26&+9.18&+5.80&\\
\multicolumn{2}{l|}{IFT on mixed stage1\&2}&-0.60&-1.61&-3.06&+0.62&+2.93&+2.71&-0.09&\\
\multicolumn{2}{l|}{IFT on mixed stage1\&3}&+1.39&+0.46&-3.33&+0.00&+6.12&+3.10&+1.33&\\
\multicolumn{2}{l|}{IFT on mixed stage2\&3}&+1.39&+3.67&+7.50&+8.12&+7.45&+10.85&+5.68&\\ \hline
\end{tabular}}
\end{table*}

Experiment 1 demonstrates the effectiveness of Phased IFT compared to One-off IFT, while Experiment 2 reveals that this effectiveness arises from multi-stage sub-datasets with increasing difficulty, not from the multi-stage training itself. Therefore, we designed multi-stage datasets with random segmentation, each segment maintaining the same quantity as the corresponding dataset with increasing difficulty, but the data points were randomly sampled. Table~\ref{tab_Exp2} lists the results of three sets of experiments as follows.

The first part of Table~\ref{tab_Exp2} presents the win rate of Phased IFT on Alpaca-rand-3-stages relative to One-off IFT on the original Alpaca. Specifically, we used five baseline models, namely Llama2 7/13/70B and Llama3 8/70B. Their performance on six benchmarks was consistently lower than that of One-off IFT, indicated by negative win rates. This suggests that segmenting the dataset randomly into multiple subsets and then performing uptraining did not yield benefits. Similarly, the second part of Table 3 shows the win rate of Phased IFT on Alpaca-cleaned-rand-3-stages relative to One-off IFT on the original Alpaca-cleaned. Based on the results, the win rates of Llama2 7B and Llama3 8B were negative, indicating that multi-stage training did not help.

The third part of Table~\ref{tab_Exp2} shows the win rate of Phased IFT on Alpaca-3-stages relative to Phased IFT on Alpaca-rand-3-stages. The results indicate that uptraining on the difficulty-increasing Alpaca-3-stages dataset resulted in significantly higher win rates than uptraining on the randomly segmented Alpaca-rand-3-stages dataset. Specifically, we used five baseline models for uptraining on both Alpaca-3-stages and Alpaca-rand-3-stages. The average win rate of Llama3 70B reached +10.62, Llama2 70B reached +9.71, Llama2 13B reached +8.29, and Llama2 7B and Llama3 8B also reached +5.74 and +5.95, respectively. These models all achieved consistently high win rates, far higher than uptraining on randomly segmented subsets.

These experiments indicate that uptraining on multi-stage datasets with increasing difficulty is beneficial and enhances the model's capabilities. Uptraining on randomly segmented multi-stage datasets reduced the model's capabilities, and the results were not as good as single-step fine-tuning on the original dataset (For a detailed graphical presentation, please refer to Appendix~\ref{sec:appendix_sec4}). This suggests that multi-stage training methods did not improve the model's capabilities and yield benefits. The effect of Phased IFT mainly comes from multi-stage subsets with increasing difficulty, which also validates our proposed progressive alignment hypothesis, that aligning pre-trained models to accomplish the final human task is a gradual process.

\subsection {Experiment 3: compares the win rates achieved by uptraining on Alpaca-3-stages across all permutations} 
To investigate the impact of varying the sequence of Alpaca-3-stages on the performance of the Llama-2 7B/13B and Llama3 8B models with clarity and precision, the following structure is proposed. This structure presents the uptraining order across the three stages, denoted as Stage 1 (1), Stage 2 (2), and Stage 3 (3), to illustrate the sequence in which the model was uptrained on different segments of the instruction data:
\begin{itemize}
\item 1-2-3: This order indicates that the model was first uptrained on data from Stage 1, followed by Stage 2, and finally Stage 3.
\item 2-1-3: The model was initially uptrained on Stage 2 data, then on Stage 1, and finally with Stage 3.
\item 3-1-2: The uptraining commenced with Stage 3 data, proceeded to Stage 1, and ended with Stage 2.
\item 1-3-2: The sequence started with Stage 1, moved to Stage 3, and finished with Stage 2.
\item 2-3-1: Uptraining began with Stage 2 data, followed by Stage 3, and then Stage 1.
\item 3-2-1: This order shows that the model was first uptrained on Stage 3 data, then Stage 2, and finally Stage 1.
\end{itemize}

Table~\ref{tab_Exp3} presents the uptraining results of Llama2 7/13B and Llama3 8B models on the three-stages Alpaca-3-stages dataset, evaluating 6 possible permutations. Analyzing the average win rate, a clear pattern emerges: the sequences 1-2-3 and 2-1-3 outperform 1-3-2 and 3-1-2, which in turn outperform 2-3-1 and 3-2-1. This suggests that uptraining on sub-datasets of increasing difficulty can indeed enhance the model's potential.

The first part of Table~\ref{tab_Exp3} details the full permutation experiments on the Llama2 7B model. The 1-2-3 sequence achieved the highest win rates across 6 benchmarks and the average win rate, except for a slight dip below the 2-1-3 sequence in the WizardLM benchmark (e.g., +6.42 vs. +6.65). The average win rate for the 1-2-3 sequence was +7.26, significantly higher than the other permutations. The 2-1-3 sequence also performed well with an average win rate of +4.59, both sequences sharing a common feature of placing the most difficult sub-dataset (stage 3) at the final stage of uptraining. The sequences 3-1-2 and 1-3-2, which place the medium difficulty sub-dataset (stage 2) last, also showed positive win rates of +2.26 and +2.14, respectively. Surprisingly, when the easiest sub-dataset (stage 1) was placed last, the sequences 2-3-1 and 3-2-1 had lower average win rates than One-off IFT on the original Alpaca dataset.

The second part of Table~\ref{tab_Exp3} shows the full permutation experiments on the Llama2 13B model. Similarly, the 1-2-3 and 2-1-3 sequences achieved the highest average win rates of +7.35 and +5.57, respectively, significantly outperforming the other permutations. These sequences also shared the characteristic of placing the most difficult sub-dataset (stage 3) last. The 1-3-2 and 3-1-2 sequences achieved average win rates of +3.53 and +2.71, respectively, with the 3-2-1 sequence having the lowest average win rate of -7.09. Interestingly, the permutation experiments on the Llama2 13B model aligned perfectly with the win rate predictions of the progressive alignment hypothesis, ranking the uptraining results from highest to lowest average win rate as follows: 1-2-3 > 2-1-3 > 1-3-2 > 3-1-2 > 2-3-1 > 3-2-1. This further confirms the benefit of uptraining on sub-datasets of increasing difficulty when fine-tuning LLMs.

The third part of Table~\ref{tab_Exp3} presents the full permutation experiments on the Llama3 8B model. The 1-2-3 and 2-1-3 sequences achieved the highest average win rates of +7.64 and +3.02, respectively, outperforming the 3-1-2 and 1-3-2 sequences, which had average win rates of +2.32 and +1.40, respectively. Unlike the Llama2 7/13B models, the 2-3-1 sequence had the lowest win rate at -13.57, significantly lower than the -8.26 of the 3-2-1 sequence. Further investigation is needed to understand the reasons behind this result.

In sum, the results indicate a phenomenon where placing more difficult sub-datasets at the final stage of uptraining on differentiated difficulty sub-datasets can significantly enhance the model's capabilities. This approach markedly increases the model's win rate, making the generated results closer to human perception, as the  judgments of GPT-4 align with human approximately 75\%.

\subsection{Ablation Studies}\label{sec:appendix_sec3}
We conducted ablation experiments using Llama2 7B/13B on Alpaca-3-stages, incorporating various sub-dataset combinations. These combinations included performing Phased IFT on Alpaca-3-stages, IFT exclusively on stage 2, IFT exclusively on stage 3, IFT on a combined sub-dataset of stage 1 and stage 2, IFT on a combined sub-dataset of stage 1 and stage 3, and IFT on a combined sub-dataset of stage 2 and stage 3. Table~\ref{tab_Exp4} presents two sets of ablation experiments, further verifying the role of each sub-dataset in fine-tuning LLMs.

The first part of Table~\ref{tab_Exp4} shows the ablation results for the Llama2 7B base model. Performing IFT exclusively on stage 2 yielded a positive gain, with a win rate of +1.14. The win rate for Phased IFT at the stage 2 phase was +3.34, indicating that stage 1 also played a positive role during uptraining. The win rate for IFT exclusively on stage 3 reached +4.44, which was higher than the Phased IFT at stage 2 but lower than the Phased IFT at stage 3. This suggests that stage 3 played a major role in performance improvement during Phased IFT. This result is consistent with the conclusions of Alpagasus \cite{Alpagasus}, which indicate that high-quality data filtered by GPT is crucial for fine-tuning models. Performing IFT on the combined sub-dataset of stage 1 and stage 2 resulted in a negative win rate, as did performing IFT on the combined sub-dataset of stage 1 and stage 3. This implies that redundant datasets are not conducive to instruction training. However, performing IFT on the combined sub-dataset of stage 2 and stage 3 resulted in a win rate of +3.53, higher than the +1.14 for IFT on stage 2 alone but lower than the +4.44 for IFT on stage 3 alone. This nuanced result suggests that mixing stage 2 data had a detrimental effect, leading to reduced model performance. Nevertheless, when uptraining across all three stages (stage 1, stage 2, and stage 3), the win rate steadily improved, indicating a gradual enhancement in model capability.

This is a very interesting experimental phenomenon, and similar observations were made in ablation experiments with Llama2 13B. This indicates that the model performance improvement achieved through uptraining on increasingly difficult sub-datasets is not a trivial training technique. The quality of the instruction data and the alignment method are key factors in fine-tuning of large language models. This also further supports the hypothesis of alignment progression as a genuine phenomenon.

\section{Conclusion}
The paper introduces the progressive alignment hypothesis, and then proposes Phased IFT approach. Utilizing GPT-4 to score the difficulty of instructions, the instruction dataset is categorized into different stages according to the difficulty. Diverging from prior methods that apply a One-off IFT on the whole instruction data, our approach entails staged, supervised uptraining on each stage in ascending order of difficulty. Through extensive experiments, we have validated the progressive alignment hypothesis and the efficacy of Phased IFT, demonstrating its utility in developing diverse instruction dataset and enhancing fine-tuning LLMs.

\section{Broader Impact}
This paper discusses a fundamental problem in Large Language Models (LLMs): How to more effectively utilize instruction datasets to align LLMs more efficiently? This paper introduces a simple and effective alignment method, which includes the analysis of instruction difficulty and structuring the dataset into multi-stage sub-datasets, with each sub-dataset increasing in difficulty. Supervised fine-tuning is then carried out through uptraining. We conducted a series of experiments to actively verify the existence of the progressive alignment hypothesis. If the progressive alignment hypothesis is a universal phenomenon, then this work will have a broad impact on the direction of fine-tuning instructions for Large Language Models.

The datasets employed in this work, namely Alpaca 52K and Alpaca-cleaned, are publicly available, and our training framework leverages open-source tools, utilizing Hugging Face's Trainer and DeepSpeed. Furthermore, we have purchased access to OpenAI's API to employ GPT-4 for assessing the difficulty of instructions and for evaluating the quality of generated responses. We make all our code and data openly available at \href{https://github.com/xubuvd/PhasedSFT}{https://github.com/xubuvd/PhasedSFT}, including training codes, GPT-4 difficulty scoring codes, win-rate calculation codes, and the difficulty-stratified Alpaca and Alpaca-cleaned instruction datasets.

\section{Limitations}
Here, we discuss some limitations of our study to inspire future research in this direction.

\noindent {\bf Automatic instruction dataset segmentation} This study draws heuristic information from the cumulative density curve of difficulty scores in the instruction dataset using human experience, selecting several thresholds to divide the instruction dataset into multiple subsets at different stages, with increasing difficulty levels. Utilizing human experience to select thresholds is a significant task that not only involves partitioning the dataset into subsets from easy to difficult but also requires considering the balance in the quantity of each subset, heavily relying on human expertise. Therefore, a significant challenge lies in developing an algorithm capable of autonomously identifying threshold points from this cumulative density curve, eliminating the dependence on human intuition, and ensuring a balanced quantity across the multi-stage subsets

\section*{Acknowledgements}
We thank the reviewers for their comments and suggestions. This paper was partially supported by the National Natural Science Foundation of China (NSFC 62076032), Huawei Noah’s Ark Lab, MoECMCC “Artificial Intelligence” Project (No. MCM20190701), Beijing Natural Science Foundation (Grant No. 4204100), and BUPT Excellent Ph.D. Students Foundation (No. CX2020309).

\bibliography{acl_latex}

\begin{thebibliography}{26}
\expandafter\ifx\csname natexlab\endcsname\relax\def\natexlab#1{#1}\fi

\bibitem[{Bai et~al.(2022)Bai, Jones, Ndousse, and et.al}]{Anthropic}
Yuntao Bai, Andy Jones, Kamal Ndousse, and et.al. 2022.
\newblock Training a helpful and harmless assistant with reinforcement learning
  from human feedback.
\newblock \emph{arXiv preprint arXiv:2204.05862}.

\bibitem[{Campello et~al.(2020)Campello, Kr{\"o}ger, Sander, and
  Zimek}]{campello2020density}
Ricardo~JGB Campello, Peer Kr{\"o}ger, J{\"o}rg Sander, and Arthur Zimek. 2020.
\newblock Density-based clustering.
\newblock \emph{Wiley Interdisciplinary Reviews: Data Mining and Knowledge
  Discovery}, 10(2):e1343.

\bibitem[{Chen et~al.(2024)Chen, Li, Yan, Wang, Gunaratna, Yadav, Tang,
  Srinivasan, Zhou, Huang, and Jin}]{Alpagasus}
Lichang Chen, Shiyang Li, Jun Yan, Hai Wang, Kalpa Gunaratna, Vikas Yadav,
  Zheng Tang, Vijay Srinivasan, Tianyi Zhou, Heng Huang, and Hongxia Jin. 2024.
\newblock Alpagasus: Training a better alpaca model with fewer data.
\newblock In \emph{ICLR}.

\bibitem[{Chia et~al.(2023)Chia, Hong, Bing, and Poria}]{chia2023instructeval}
Yew~Ken Chia, Pengfei Hong, Lidong Bing, and Soujanya Poria. 2023.
\newblock Instructeval: Towards holistic evaluation of instruction-tuned large
  language models.
\newblock \emph{arXiv preprint arXiv:2306.04757}.

\bibitem[{Chiang et~al.(2023)Chiang, Li, Lin, Sheng, Wu, Zhang, Zheng, Zhuang,
  Zhuang, Gonzalez, Stoica, and Xing}]{Vicuna}
Wei-Lin Chiang, Zhuohan Li, Zi~Lin, Ying Sheng, Zhanghao Wu, Hao Zhang, Lianmin
  Zheng, Siyuan Zhuang, Yonghao Zhuang, Joseph~E. Gonzalez, Ion Stoica, and
  Eric~P. Xing. 2023.
\newblock \href {https://lmsys.org/blog/2023-03-30-vicuna/} {Vicuna: An
  open-source chatbot impressing gpt-4 with 90\%* chatgpt quality}.

\bibitem[{Dubois et~al.(2023)Dubois, Li, Taori, Zhang, Gulrajani, Ba, Guestrin,
  Liang, and Hashimoto}]{AlpacaFarm}
Yann Dubois, Xuechen Li, Rohan Taori, Tianyi Zhang, Ishaan Gulrajani, Jimmy Ba,
  Carlos Guestrin, Percy Liang, and Tatsunori~B. Hashimoto. 2023.
\newblock Alpacafarm: A simulation framework for methods that learn from human
  feedback.
\newblock In \emph{NeurIPS}.

\bibitem[{Geng et~al.(2023)Geng, Gudibande, Liu, Wallace, Abbeel, Levine, and
  Song}]{Koala}
Xinyang Geng, Arnav Gudibande, Hao Liu, Eric Wallace, Pieter Abbeel, Sergey
  Levine, and Dawn Song. 2023.
\newblock \href {https://bair.berkeley.edu/blog/2023/04/03/koala/} {Koala: A
  dialogue model for academic research}.
\newblock Blog post.

\bibitem[{Jiang et~al.(2023)Jiang, Sablayrolles, Mensch, Bamford, Chaplot,
  de~las Casas, Bressand, Lengyel, Lample, Saulnier, Lavaud, Lachaux, Stock,
  Scao, Lavril, Wang, Lacroix, and Sayed}]{Mistral7B}
Albert~Q. Jiang, Alexandre Sablayrolles, Arthur Mensch, Chris Bamford,
  Devendra~Singh Chaplot, Diego de~las Casas, Florian Bressand, Gianna Lengyel,
  Guillaume Lample, Lucile Saulnier, L\'elio~Renard Lavaud, Marie-Anne Lachaux,
  Pierre Stock, Teven~Le Scao, Thibaut Lavril, Thomas Wang, Timoth\'ee Lacroix,
  and William~El Sayed. 2023.
\newblock Mistral 7b.
\newblock \emph{arXiv preprint arXiv:2310.06825}.

\bibitem[{K{\"o}pf et~al.(2023)K{\"o}pf, Kilcher, von R{\"u}tte, Anagnostidis,
  Tam, Stevens, Barhoum, Duc, Stanley, Nagyfi et~al.}]{kopf2023openassistant}
Andreas K{\"o}pf, Yannic Kilcher, Dimitri von R{\"u}tte, Sotiris Anagnostidis,
  Zhi-Rui Tam, Keith Stevens, Abdullah Barhoum, Nguyen~Minh Duc, Oliver
  Stanley, Rich{\'a}rd Nagyfi, et~al. 2023.
\newblock Openassistant conversations--democratizing large language model
  alignment.
\newblock \emph{arXiv preprint arXiv:2304.07327}.

\bibitem[{Li et~al.(2024)Li, Zhang, He, Li, Zhao, Wang, Cheng, and
  Zhou}]{li2024superfiltering}
Ming Li, Yong Zhang, Shwai He, Zhitao Li, Hongyu Zhao, Jianzong Wang, Ning
  Cheng, and Tianyi Zhou. 2024.
\newblock \href {http://arxiv.org/abs/2402.00530} {Superfiltering:
  Weak-to-strong data filtering for fast instruction-tuning}.

\bibitem[{Li et~al.(2023{\natexlab{a}})Li, Zhang, Li, Chen, Chen, Cheng, Wang,
  Zhou, and Xiao}]{Quantity}
Ming Li, Yong Zhang, Zhitao Li, Jiuhai Chen, Lichang Chen, Ning Cheng, Jianzong
  Wang, Tianyi Zhou, and Jing Xiao. 2023{\natexlab{a}}.
\newblock \href {http://arxiv.org/abs/2308.12032} {From quantity to quality:
  Boosting llm performance with self-guided data selection for instruction
  tuning}.

\bibitem[{Li et~al.(2023{\natexlab{b}})Li, Ma, Wang, Huang, Jiang, Zheng, Xie,
  Huang, and Jiang}]{EcomGPT}
Yangning Li, Shirong Ma, Xiaobin Wang, Shen Huang, Chengyue Jiang, Hai-Tao
  Zheng, Pengjun Xie, Fei Huang, and Yong Jiang. 2023{\natexlab{b}}.
\newblock Ecomgpt: Instruction-tuning large language models with chain-of-task
  tasks for e-commerce.
\newblock \emph{arXiv:2308.06966}.

\bibitem[{Longpre et~al.(2023)Longpre, Hou, Vu, Webson, Chung, Tay, Zhou, Le,
  Zoph, Wei, and Roberts}]{Flan}
Shayne Longpre, Le~Hou, Tu~Vu, Albert Webson, Hyung~Won Chung, Yi~Tay, Denny
  Zhou, Quoc~V. Le, Barret Zoph, Jason Wei, and Adam Roberts. 2023.
\newblock The flan collection: designing data and methods for effective
  instruction tuning.
\newblock In \emph{ICML}.

\bibitem[{OpenAI(2023)}]{GPT-4}
OpenAI. 2023.
\newblock Gpt-4 technical report.
\newblock \emph{arXiv preprint arXiv:2303.08774}.

\bibitem[{Ouyang et~al.(2022)Ouyang, Wu, Jiang, Almeida, Wainwright, and
  etc.}]{InstructGPT}
Long Ouyang, Jeff Wu, Xu~Jiang, Diogo Almeida, Carroll~L. Wainwright, and etc.
  2022.
\newblock Training language models to follow instructions with human feedback.
\newblock In \emph{NeurIPS}.

\bibitem[{Taori et~al.(2023)Taori, Gulrajani, Zhang, Dubois, Li, Guestrin,
  Liang, and Hashimoto}]{Alpaca}
Rohan Taori, Ishaan Gulrajani, Tianyi Zhang, Yann Dubois, Xuechen Li, Carlos
  Guestrin, Percy Liang, and Tatsunori~B Hashimoto. 2023.
\newblock Alpaca: A strong, replicable instruction-following model.
\newblock \emph{Stanford Center for Research on Foundation Models}, 3.

\bibitem[{Thoppilan et~al.(2022)Thoppilan, Freitas, Hall, Shazeer, and
  et.al}]{LaMDA}
Romal Thoppilan, Daniel~De Freitas, Jamie Hall, Noam Shazeer, and et.al. 2022.
\newblock Lamda: Language models for dialog applications.
\newblock \emph{arXiv preprint arXiv:2201.08239}.

\bibitem[{Touvron et~al.(2023)Touvron, Martin, and et.al}]{LLaMA2}
Hugo Touvron, Louis Martin, and et.al. 2023.
\newblock Llama 2: Open foundation and fine-tuned chat models.
\newblock \emph{arXiv preprint arXiv:2307.09288}.

\bibitem[{Wang et~al.(2023{\natexlab{a}})Wang, Li, Chen, Cai, Zhu, Lin, Cao,
  Liu, Liu, and Sui}]{Fair}
Peiyi Wang, Lei Li, Liang Chen, Zefan Cai, Dawei Zhu, Binghuai Lin, Yunbo Cao,
  Qi~Liu, Tianyu Liu, and Zhifang Sui. 2023{\natexlab{a}}.
\newblock Large language models are not fair evaluators.
\newblock \emph{arXiv preprint arXiv:2305.17926}.

\bibitem[{Wang et~al.(2023{\natexlab{b}})Wang, Kordi, Mishra, Liu, Smith,
  Khashabi, and Hajishirzi}]{SELF-INSTRUCT}
Yizhong Wang, Yeganeh Kordi, Swaroop Mishra, Alisa Liu, Noah~A. Smith, Daniel
  Khashabi, and Hannaneh Hajishirzi. 2023{\natexlab{b}}.
\newblock Self-instruct: Aligning language models with self-generated
  instructions.
\newblock In \emph{ACL}.

\bibitem[{Wei et~al.(2022)Wei, Wang, Schuurmans, Bosma, Ichter, Xia, Quoc, and
  Zhou}]{CoT}
Jason Wei, Xuezhi Wang, Dale Schuurmans, Maarten Bosma, Brian Ichter, Fei Xia,
  Ed~H.~Chi Quoc, and V.~Le~Denny Zhou. 2022.
\newblock Chain-of-thought prompting elicits reasoning in large language
  models.
\newblock In \emph{NeurIPS}.

\bibitem[{Xu et~al.(2024)Xu, Sun, Zheng, Geng, Zhao, Feng, Tao, Lin, and
  Jiang}]{WizardLM}
Can Xu, Qingfeng Sun, Kai Zheng, Xiubo Geng, Pu~Zhao, Jiazhan Feng, Chongyang
  Tao, Qingwei Lin, and Daxin Jiang. 2024.
\newblock Wizardlm: Empowering large pre-trained language models to follow
  complex instructions.
\newblock In \emph{ICLR}.

\bibitem[{yahma(2023)}]{alpaca-cleaned}
yahma. 2023.
\newblock {yahma/alpaca-cleaned}.
\newblock \url{https://huggingface.co/datasets/yahma/alpaca-cleaned}.
\newblock [Online; accessed 11-October-2023].

\bibitem[{Zhao et~al.(2023)Zhao, Zhou, Li, Tang, Wang, Hou, Min, Zhang, Zhang,
  Dong et~al.}]{zhao2023survey}
Wayne~Xin Zhao, Kun Zhou, Junyi Li, Tianyi Tang, Xiaolei Wang, Yupeng Hou,
  Yingqian Min, Beichen Zhang, Junjie Zhang, Zican Dong, et~al. 2023.
\newblock A survey of large language models.
\newblock \emph{arXiv preprint arXiv:2303.18223}.

\bibitem[{Zheng et~al.(2023)Zheng, Chiang, Sheng, Zhuang, Wu, Zhuang, Lin, Li,
  Li, Xing et~al.}]{zheng2023judging}
Lianmin Zheng, Wei-Lin Chiang, Ying Sheng, Siyuan Zhuang, Zhanghao Wu, Yonghao
  Zhuang, Zi~Lin, Zhuohan Li, Dacheng Li, Eric Xing, et~al. 2023.
\newblock Judging llm-as-a-judge with mt-bench and chatbot arena.
\newblock \emph{arXiv preprint arXiv:2306.05685}.

\bibitem[{Zhou et~al.(2023)Zhou, Liu, Xu, Iyer, Sun, Mao, Ma, Efrat, Yu, Yu,
  Zhang, Ghosh, Lewis, Zettlemoyer, and Levy}]{LIMA}
Chunting Zhou, Pengfei Liu, Puxin Xu, Srini Iyer, Jiao Sun, Yuning Mao, Xuezhe
  Ma, Avia Efrat, Ping Yu, Lili Yu, Susan Zhang, Gargi Ghosh, Mike Lewis, Luke
  Zettlemoyer, and Omer Levy. 2023.
\newblock Lima: Less is more for alignment.
\newblock In \emph{NeurIPS}.

\end{thebibliography}

\appendix

\section{Appendix} \label{sec:appendix}

\begin{table}[!htb]
\centering 
\caption{Comparison of percentage distribution of difficulty scores on the Alpaca52K dataset with GPT-3.5 and GPT-4} \label{tab_Exp_score_dist}
\setlength{\tabcolsep}{1.0mm}{
\begin{tabular}{lcccc}
\hline 
GPT& {[1.0, 2.5)} &{[2.5, 3.5]}& {(3.5, 4.5)} & {[4.5, 5.0]} \\ \cline{1-5}
3.5&2.18\%&62.28\%&25.96\%&9.56\%\\
4&76.3\%&13.38\%&4.1\%&6.17\%\\
\hline
\end{tabular}}
\end{table}
\begin{figure*}[!htb]
 \includegraphics[width=0.49\linewidth,scale=0.5,clip=true]{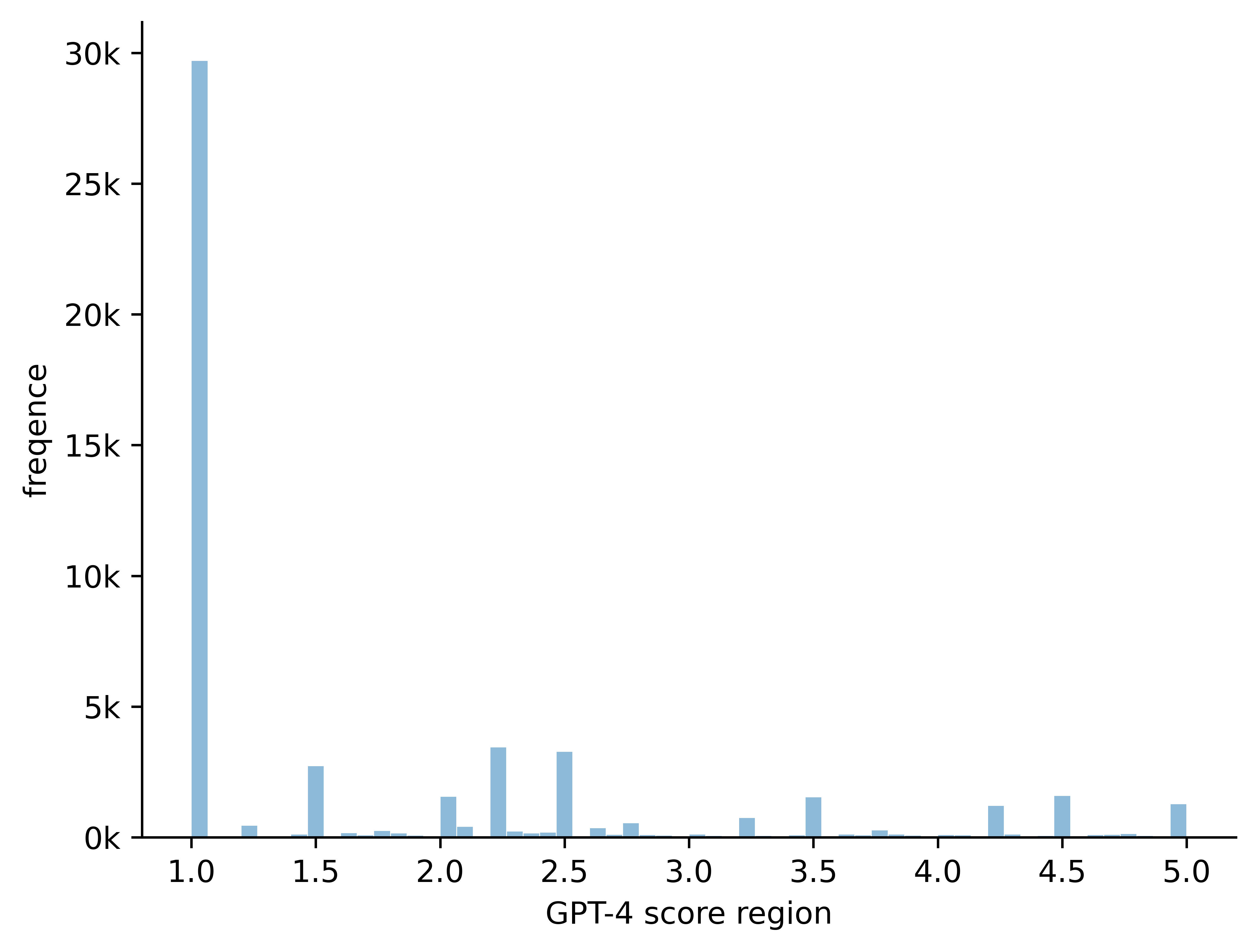}
 \hspace{0in}
 \includegraphics[width=0.49\linewidth,scale=0.5,clip=true]{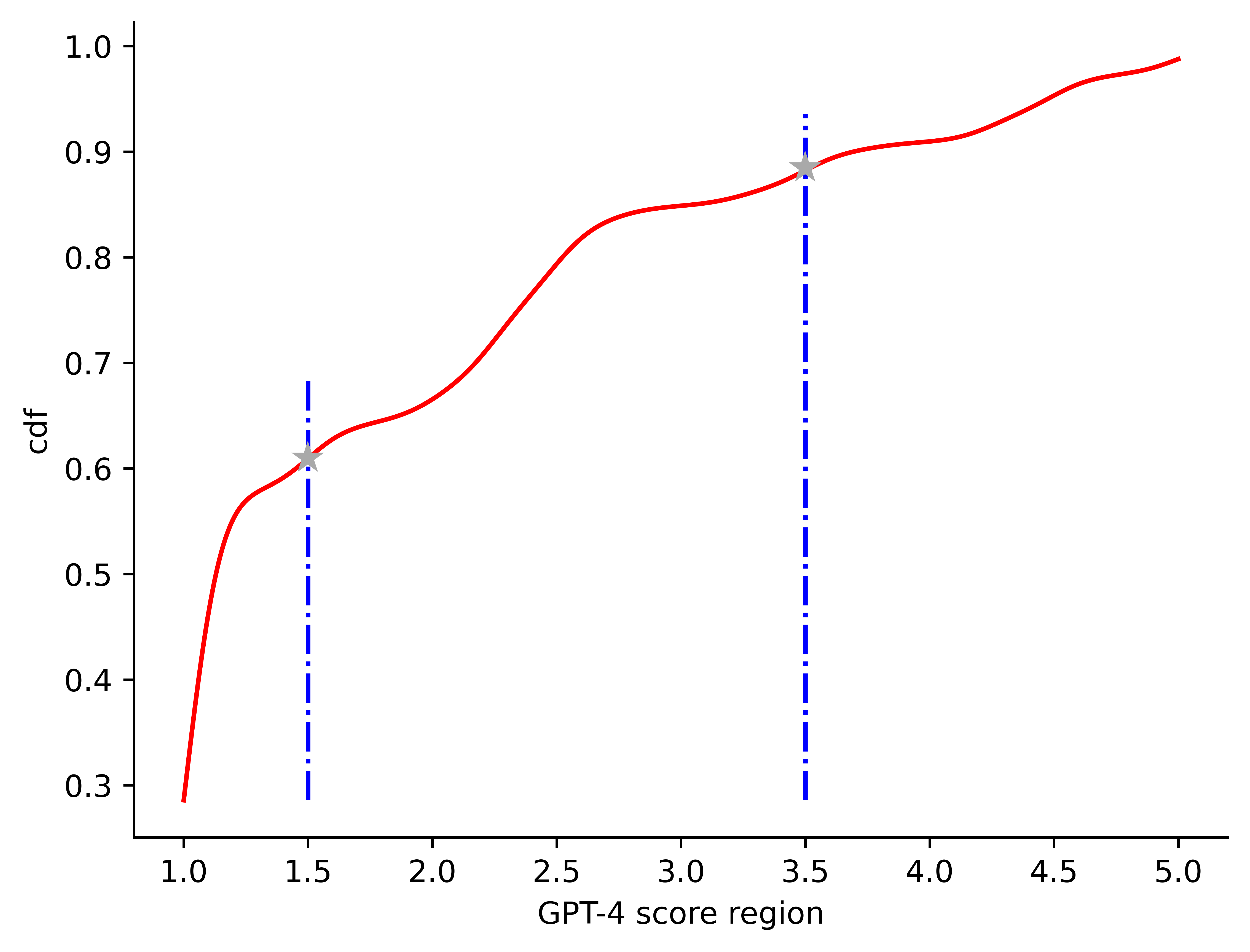}
\caption{Histogram and cumulative probability density of difficulty scores for Alpaca 52K dataset.} \label{fig_appendix_Alpaca}
\end{figure*}
\begin{figure*}[!htb]
 \includegraphics[width=0.49\linewidth,scale=0.5,clip=true]{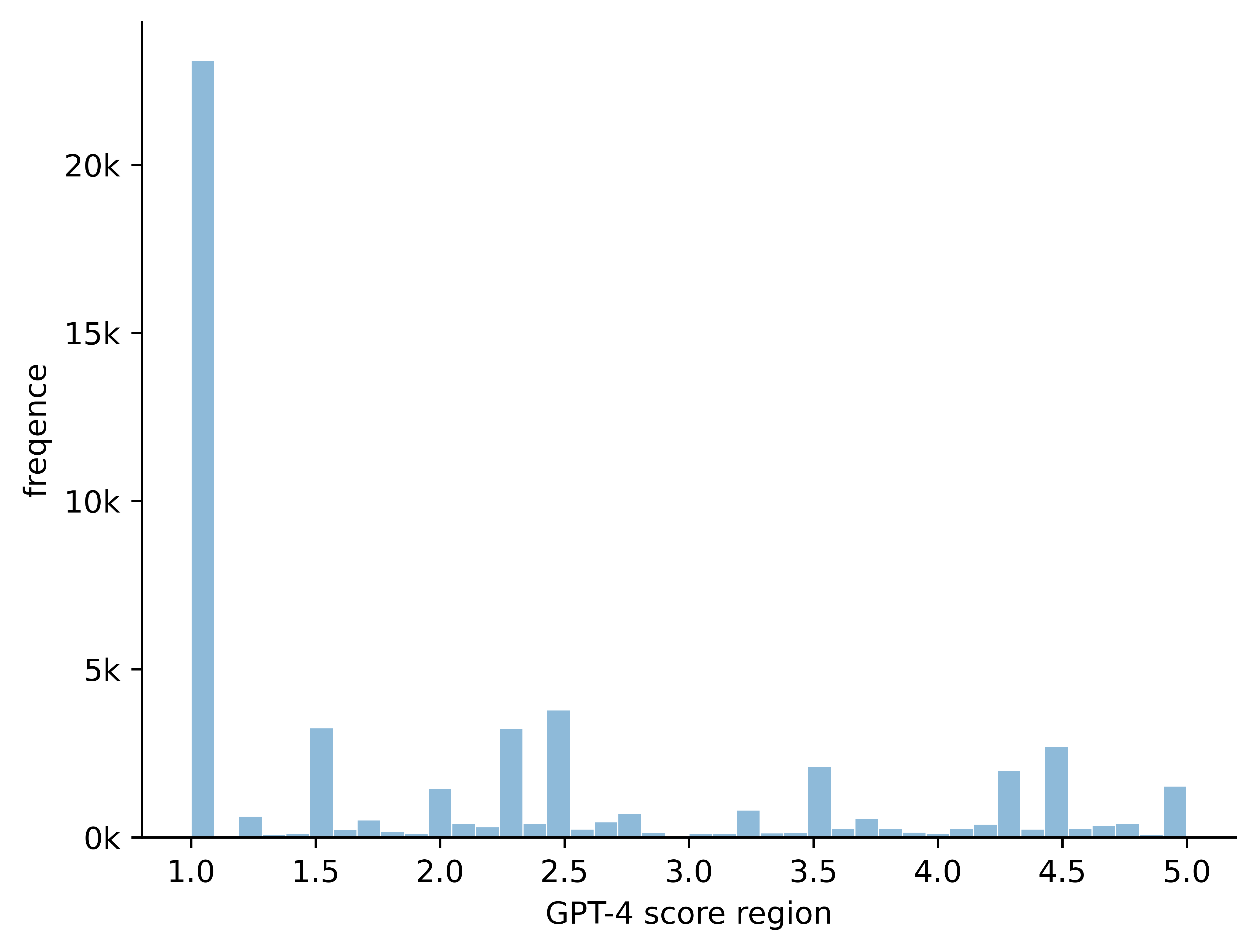}
 \hspace{0in}
 \includegraphics[width=0.49\linewidth,scale=0.5,clip=true]{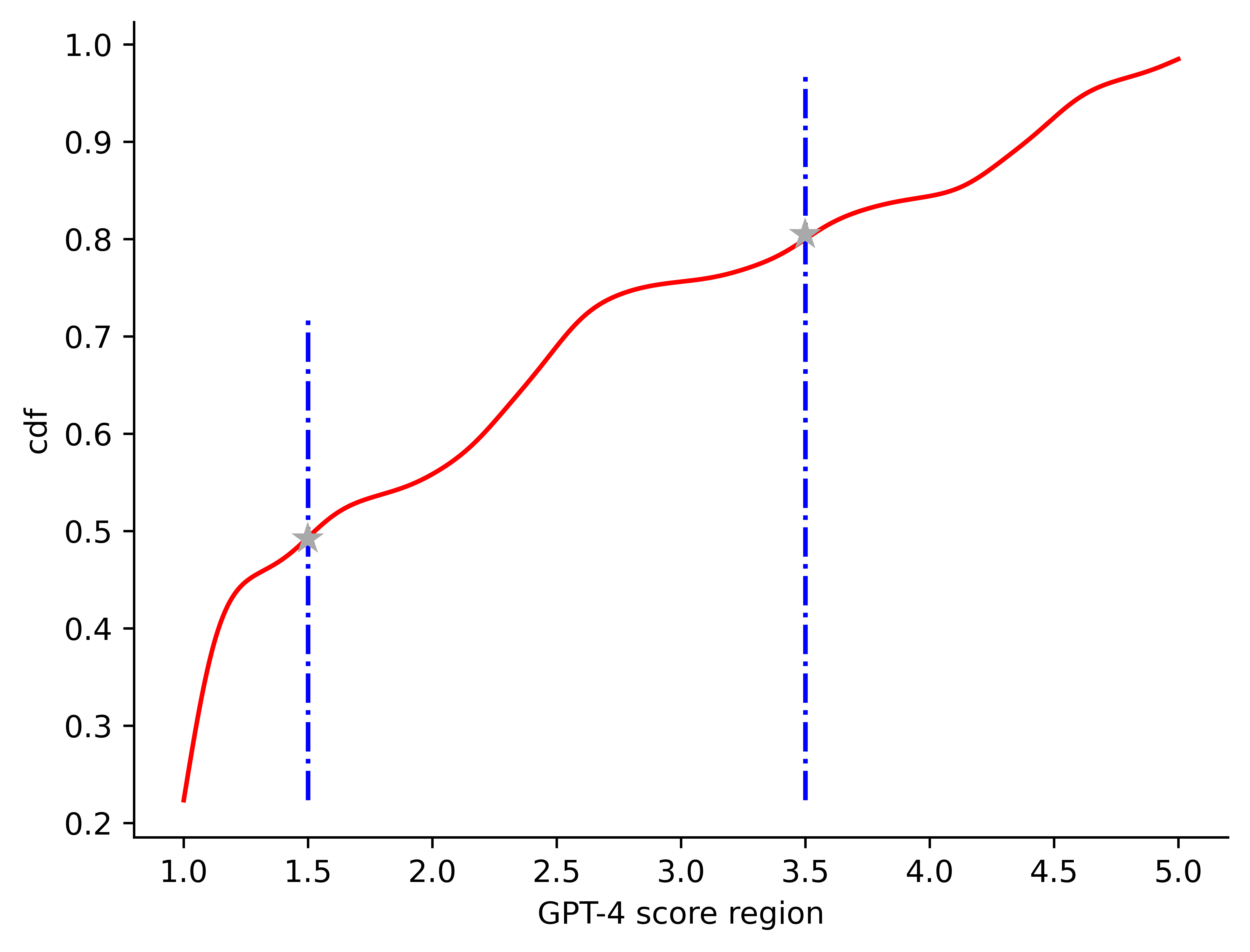}
\caption{Histogram and cumulative probability density of difficulty scores for Alpaca-cleaned 52K dataset.} \label{fig_appendix_AlpacaCleaned}
\end{figure*}
\begin{figure*}[!htb]
 \includegraphics[width=0.49\linewidth,scale=0.5,clip=true]{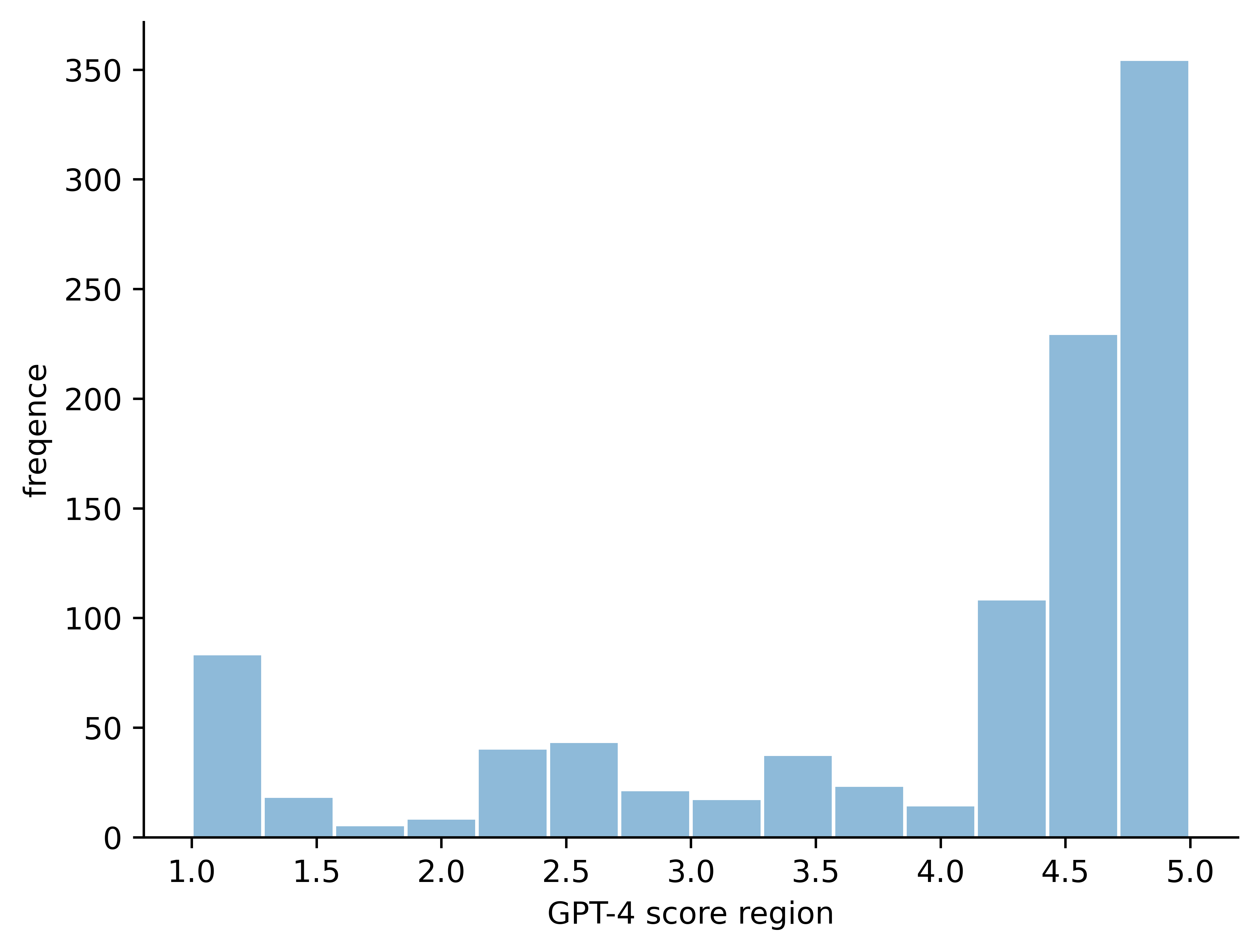}
 \hspace{0in}
 \includegraphics[width=0.49\linewidth,scale=0.5,clip=true]{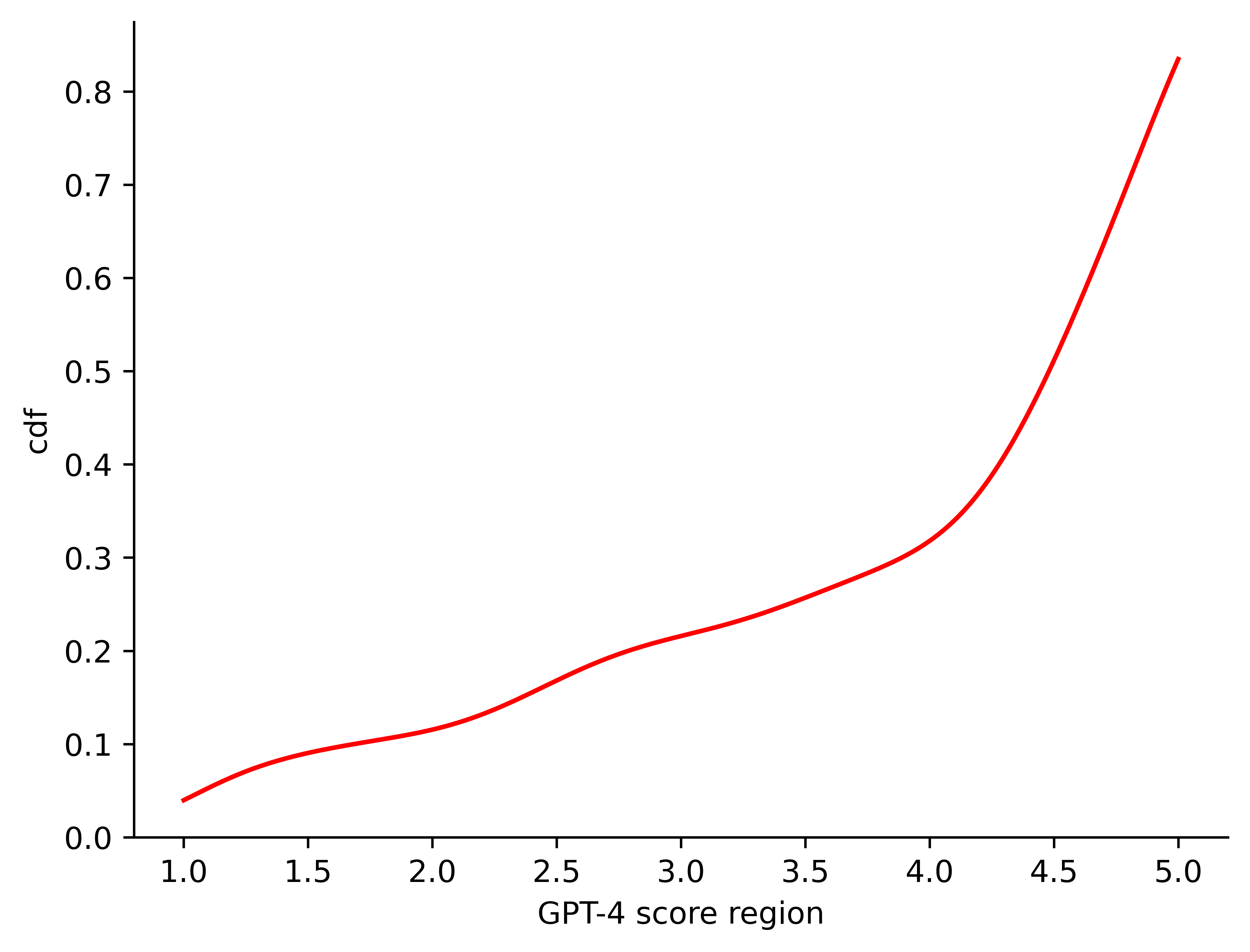}
\caption{Histogram and cumulative probability density of difficulty scores for LIMA 1000 dataset.} \label{fig_appendix_LIMA}
\end{figure*}
\begin{figure*}[!htb]
\centering 
\includegraphics[width=0.95\linewidth,scale=0.5, clip=true, keepaspectratio]{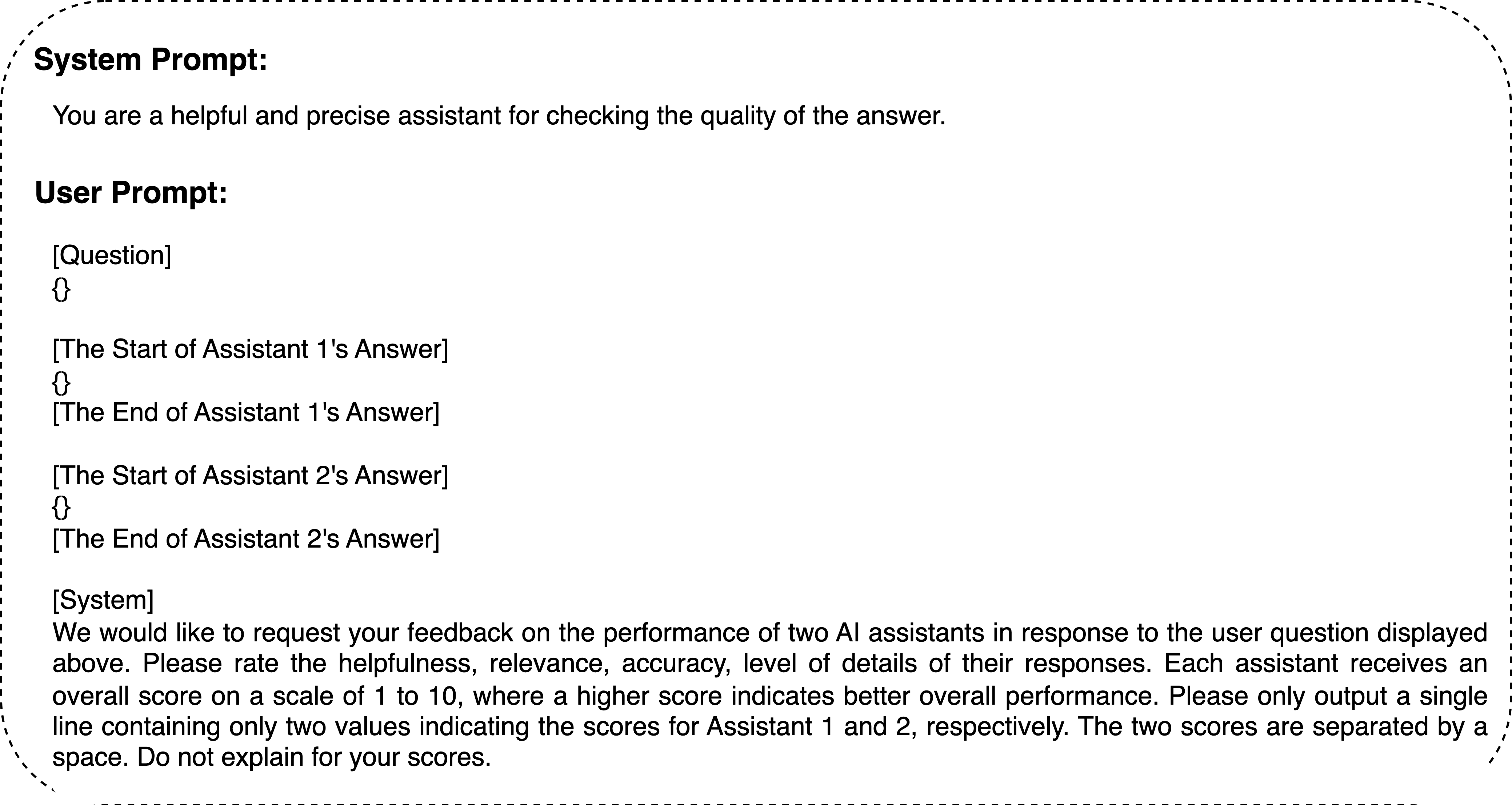}
\caption{A prompt to ChatGPT-4 for rating model's responses with a score between 1 and 10.}
\label{fig_appendix_prompt}
\end{figure*}
\begin{figure*}[!htb]
\centering 
\includegraphics[width=0.95\linewidth,scale=0.5, clip=true, keepaspectratio]{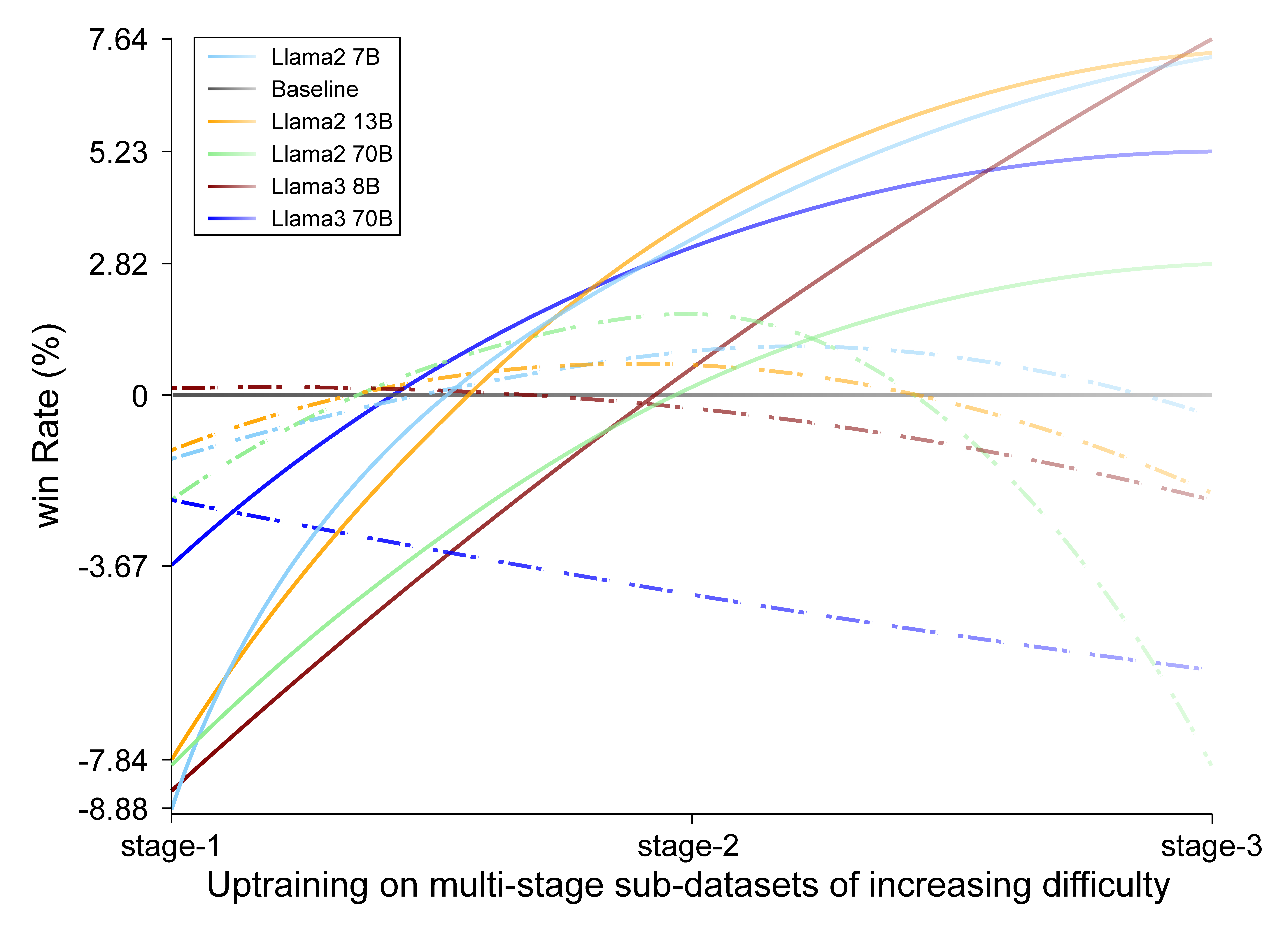}
\caption{Comparison of the win rate trends for uptraining on multi-stage sub-datasets with increasing difficulty (solid line) and uptraining on randomly divided multi-stage sub-datasets (dashed line). The gray horizontal line in the middle represents the baseline of One-off IFT on the original dataset.}
\label{fig_appendix_main}
\end{figure*}

\subsection{Comparison of percentage distribution of difficulty scores on the Alpaca52K dataset using GPT-3.5 and GPT-4} \label{sec:appendix_sec3}

In our initial assessment, we compared the differences of GPT-3.5 and GPT-4 (gpt-4-0613) in assigning difficulty scores to the Alpaca 52K dataset through human evaluation, the score ranges from 1.0 to 5.0. We divided the scores between 1.0 and 5.0 into four intervals, with the quantity percentage of each interval shown in the Table~\ref{tab_Exp_score_dist}.

The distribution of scores reveal a noticeable regression toward the mean for GPT-3.5, with scores between 2.5 and 3.5 accounting for 62.28\% of the total. A manual review of 200 samples indicated a mismatch between the assigned difficulty levels and human judgment, particularly noting an excess of simpler questions within the dataset. In contrast, GPT-4's difficulty ratings aligned with human evaluations for approximately 75\% of the cases, with scores ranging between 2.5 and 3.5 comprising 13.38\% of the total.

\subsection{Histogram and density curve for Alpaca, Alpaca-cleaned and LIMA} \label{sec:appendix_sec1}

 \noindent We conducted a comparative analysis on the instruction difficulty of three datasets—Alpaca, Alpaca-cleaned, and LIMA—employing GPT-4 for scoring. Figures~\ref{fig_appendix_Alpaca}, \ref{fig_appendix_AlpacaCleaned} to \ref{fig_appendix_LIMA} illustrate their respective score distributions and probability density functions. Alpaca, generated by the Self-Instruct framework, consists of 52K triplets (instruction, input, output). Alpaca-cleaned represents an enhanced version of Alpaca, refined to minimize errors such as incorrect responses and hallucinations. LIMA, by contrast, is composed of 1,000 manually curated instructions. Analysis of score histograms across these figures indicates a predominance of higher scores in LIMA, comparatively fewer in Alpaca, with Alpaca-cleaned exhibiting overall improved scoring relative to Alpaca. These findings corroborate our qualitative assessment of the datasets and affirm the utility of GPT-4 in evaluating instruction difficulty.

The following list includes seven examples of instructions scored by GPT-4, with respective scores of 1, 1.5, 2.6, 3.25, 4.5, 4.75, and 5. The scoring prompt is provided in Figure~\ref{fig_prompt}.
\begin{itemize}
\item {\bfseries Score: 1.0}, {\bfseries Instruction:} Given the following definition, build a sentence using the word. {\bfseries Input:} To send away abruptly or angrily. {\bfseries Output:} He angrily dismissed me without even giving me an explanation.
\item {\bfseries Score: 1.5}, {\bfseries Instruction:} Generate two points that could be used to extend the following statement. {\bfseries Input:} Internet has become an important way to access knowledge. {\bfseries Output:} - It has also become a key tool for collaboration and communication. - It has provided a platform for the sharing of ideas and greater access to global markets.
\item {\bfseries Score: 2.6}, {\bfseries Instruction:} Propose two appropriate and different methods to solve the given problem. {\bfseries Input:} Predicting whether an email is a spam or not. {\bfseries Output:} Two possible methods for predicting whether an email is a spam or not are supervised machine learning models such as artificial neural networks and Naive Bayes classifiers, or an unsupervised machine learning method such as clustering and anomaly detection.
\item {\bfseries Score: 3.25}, {\bfseries Instruction:} Edit this passage to make it more engaging to readers. {\bfseries Input:} We went for a walk in the park and saw a lot of trees. {\bfseries Output:} As we strolled through the crisp park, we were treated to sights of lush green trees rustling in the wind, stretching for miles with their towering dignity.
\item {\bfseries Score: 4.5}, {\bfseries Instruction:} What is a 9-digit number that is divisible by 7? {\bfseries Input:}. {\bfseries Output:} 105,714,285 is a 9-digit number that is divisible by 7.
\item {\bfseries Score: 4.75}, {\bfseries Instruction:} Generate a sentence that has an implied metaphor. {\bfseries Input:}. {\bfseries Output:} My love for her was a warm summer breeze, picking me up and carrying me away.
\item {\bfseries Score: 5}, {\bfseries Instruction:} Design a logo for a restaurant called 'Taste of Home'.{\bfseries Input:}. {\bfseries Output:} A font-based logo in a handwritten script style, with the words 'Taste of Home' in an orange color for a warm and inviting feeling. The logo should also contain an image of a fork and spoon in the same color.
\end{itemize}

\subsection{ChatGPT-4 for evaluating win rate} \label{sec:appendix_sec2}

Figure~\ref{fig_appendix_prompt} details the prompt used for scoring and comparing the quality of output generated by two models using GPT-4. To mitigate any potential positional bias \cite{Fair, Alpagasus} from GPT-4, we conduct two scoring sessions. Specifically, we provide GPT-4 with inputs in two different orders: (instruction, input, output-1 of model 1, output-2 of model 2) and (instruction, input, output-2 of model 2, output-1 of model 1) for separate evaluations.

\subsection{Comparison of win rate trends across multi-stages of uptraining} \label{sec:appendix_sec4}

Figure~\ref{fig_appendix_main} is a summary of this paper, which incorporates the results from Table~\ref{tab_Exp2}. As can be seen from the figure, with the progression of uptraining, it demonstrates the winning rate growth trend of five LLMs on multi-stage sub-datasets with increasing difficulty. This forms a stark contrast to the winning rate trend of the same five LLMs on multi-stage sub-datasets with randomly distributed difficulty levels.

To be specific, the winning rate steadily continues to increase when uptraining on the sub-datasets with incremental difficulty. Conversely, when uptraining on the sub-datasets with random difficulty, the winning rate first increases then decreases, apart from the continuously declining winning rate trend of Llama3 70B. This indicates that the Progressive Alignment Hypothesis we proposed is an effective instruction learning method. It gradually learns from the pre-training model's ability to predict the next word to eventually acquire the ability to complete the human end task.

\end{document}